\documentclass[10pt, a4paper]{article}

\usepackage{lrec-coling2024} 
\usepackage{graphicx}
\usepackage{tabularx}
\usepackage{soul}
\usepackage{tikz}
\usepackage{pgfplots}
\usepackage{booktabs}
\usepackage{tabularx}
\usepackage{subcaption}
\usepackage{newfloat}
\usepackage{listings}
\usepackage{graphicx}
\usepackage{algorithm}
\usepackage{algorithmic}
\usepackage{amsmath,bm}
\usepackage{pifont}
\usepackage{array}
\usepackage{multirow}
\usepackage{graphicx}
\usepackage{amssymb}
\usepackage{booktabs}
\title{ChatASU: Evoking LLM’s Reflexion to Truly Understand Aspect Sentiment in Dialogues}
\name{Yiding Liu, Jingjing Wang$^*$\thanks{\@ \@  $^*$Corresponding Author: Jingjing Wang.}, Jiamin Luo, Tao Zeng, Guodong Zhou} 

\address{School of Computer Science and Technology, Soochow University, China \\
         No.1, Shizi Street, Suzhou City, Jiangsu Province, China \\
         \{20224227068, 20204027003, 20215227014\}@stu.suda.edu.cn, \{djingwang, gdzhou\}@suda.edu.cn\\}

\abstract{
Aspect Sentiment Understanding (ASU) in interactive scenarios (e.g., Question-Answering and Dialogue) has attracted ever-more interest in recent years and achieved important progresses. However, existing studies on interactive ASU largely ignore the coreference issue for opinion targets (i.e., aspects), while this phenomenon is ubiquitous in interactive scenarios especially dialogues, limiting the ASU performance. Recently, large language models (LLMs) shows the powerful ability to integrate various NLP tasks with the chat paradigm. In this way, this paper proposes a new \textbf{Chat}-based \textbf{A}spect \textbf{S}entiment \textbf{U}nderstanding (ChatASU) task, aiming to explore LLMs' ability in understanding aspect sentiments in dialogue scenarios. Particularly, this ChatASU task introduces a sub-task, i.e., Aspect Chain Reasoning (ACR) task, to address the aspect coreference issue. On this basis, we propose a \textbf{T}rusted \textbf{S}elf-reflexion \textbf{A}pproach (TSA) with ChatGLM as backbone to ChatASU. Specifically, this TSA treats the ACR task as an auxiliary task to boost the performance of the primary ASU task, and further integrates trusted learning into reflexion mechanisms to alleviate the LLMs-intrinsic factual hallucination problem in TSA. Furthermore, a high-quality ChatASU dataset is annotated to evaluate TSA, and extensive experiments show that our proposed TSA can significantly outperform several state-of-the-art baselines, justifying the effectiveness of TSA to ChatASU and the importance of considering the coreference and hallucination issues in ChatASU.
\\ \newline }

\begin{document}

\maketitleabstract

\section{Introduction}
Aspect Sentiment Understanding (ASU), a fine-grained sentiment analysis task in the field of sentiment analysis~\cite{dingyi1,dingyi2}, centers on the extraction of aspects from individual sentences and the subsequent prediction of their sentiment polarity~\cite{juzijiABSA,qa}. Throughout the last decade, ASU has garnered widespread utilization across diverse fields, exemplified by its application in e-commerce customer service~\cite{shangwu} and social opinion mining~\cite{yijian}. Recently, some studies focus on the aspect of interactive scenarios, encompassing both single-turn Question-Answering~\cite{wangg} and multi-turn dialogue~\cite{casa}. 

\begin{figure}
\setlength{\abovecaptionskip}{0.5 ex}
\setlength{\belowcaptionskip}{-5 ex}
\begin{center}
    \subfloat{
 \includegraphics[scale=0.095]{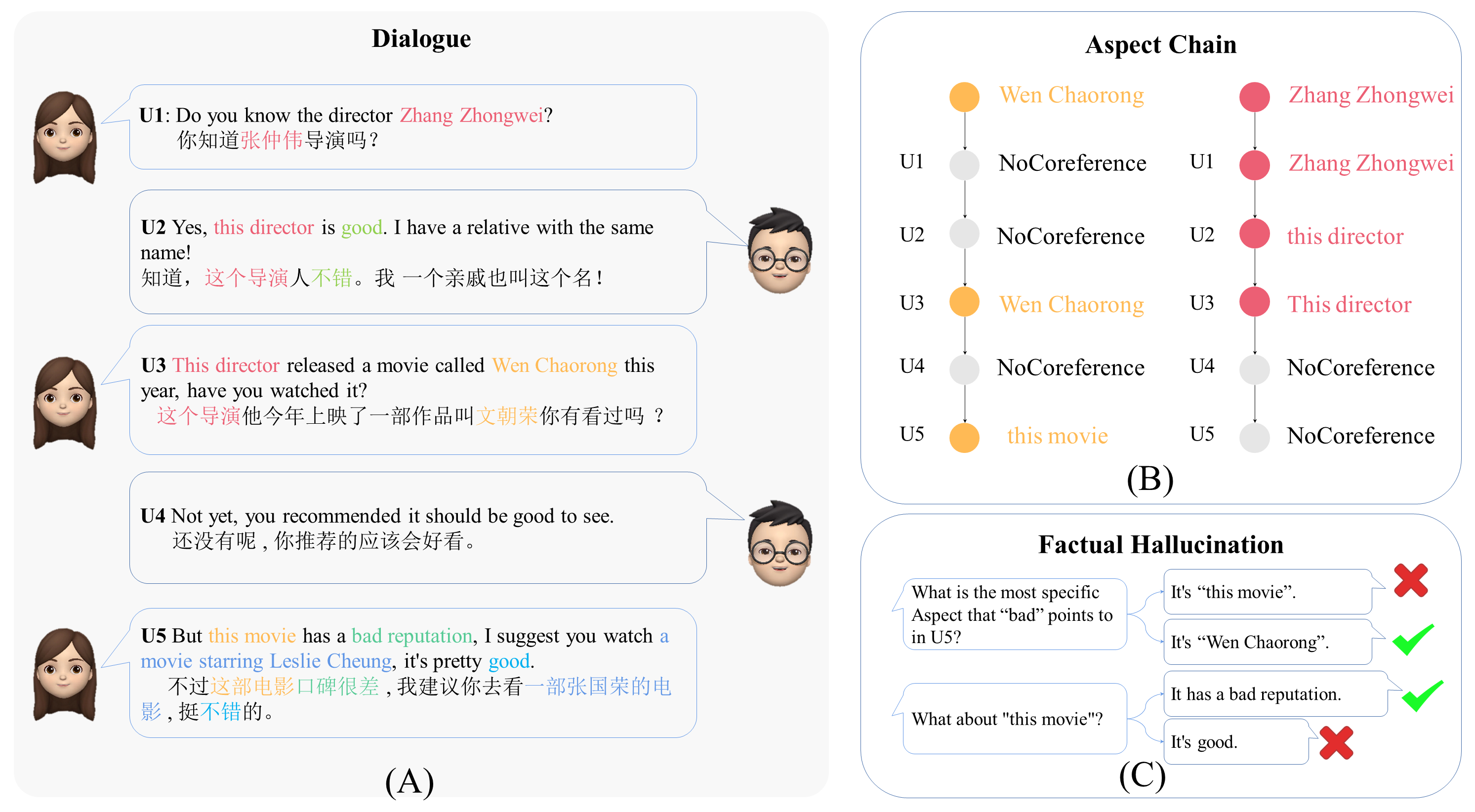}}
\caption{An example to illustrate the coreference and hallucination issue. (A): The concrete dialogue to explain the proposed Aspect Chain and Hallucinations in ChatASU, where different colors represent different aspects. (B): Two Aspect Chains of ``Wen Chaorong'' and ``Zhang Zhongwei'' with there corresponding coreference in this dialogue, where NoCoreference means that the current utterance has no coreference. (C): The factual hallucinations exist in ChatASU, i.e., errors in extracting the coreference and predicting the sentiment.} 
\label{fig:introduction}
\end{center}
\end{figure}

Despite the important progresses achieved by existing studies in ASU, they remain confined to the pre-trained language models (PLMs) phase~\cite{useBert} and ignore the coreference issue under interactive ASU scenarios. The advent and rapid advancements of large language models (LLMs) like ChatGPT shows the powerful ability to integrate various NLP tasks with the chat paradigm~\cite{llmyingyong}. Therefore, to better evaluate the ability of LLMs in understanding aspect sentiments under dialogue scenarios, we propose a new \textbf{Chat} \textbf{A}spect \textbf{S}entiment \textbf{U}nderstanding (ChatASU) task and meticulously annotate a high-quality ChatASU dataset (see details in Section \ref{sec:dataset}). In this paper, we believe that our ChatASU task faces two major challenges.


For one thing, how to address the coreference issue for aspects (namely aspect chain issue for short) in dialogues is challenging for LLMs, which could assist in precisely predicting the aspect sentiments. As shown in Figure \ref{fig:introduction} (B), we can see the aspect chain (``\emph{Wen Chaorong} $\rightarrow$ ... $\rightarrow$\emph{this movie}'') within utterances. The aspect ``\emph{Wen Chaorong}'' is not referred in utterances \textbf{U1}, \textbf{U2} and \textbf{U4} (i.e., noconference), but appears in \textbf{U3} and \textbf{U5} along with ``\emph{this movie}'' (i.e., coreference). Thus, we denote the aspect ``\emph{Wen Chaorong}'' in \textbf{U3} as \emph{Explicit} aspect and ``\emph{this movie}'' in \textbf{U5} as \emph{Implicit} aspect, where the sentiment \emph{bad} only appears in \textbf{U5}, leading to the difficulties of mapping sentiments to aspects. Therefore, a well-behaved approach should consider aspect chain to address the coreference issue and enhance LLMs' ability of understanding aspect sentiments in dialogue scenarios.

For another, LLMs usually exhibit factual hallucination problem in their generative and predictive capabilities~\cite{huanjue1,huanjue2}. More seriously, due to the existence of aspect chain in ChatASU, LLMs face more serious factual hallucination challenges. Also as shown in Figure \ref{fig:introduction} (C), a right instance of aspect chain involving ``\emph{Wen Chaorong}'' is associated with ``\emph{this movie}''. However, models often misunderstand the context and erroneously link ``\emph{this movie}'' to ``\emph{Zhang Zhongwei}'', resulting in the coreference error of factual hallucination issues. Moreover, for the implicit aspect ``\emph{this movie}'', LLMs tend to predict \emph{good} (in \textbf{U2}) instead of \emph{bad} (in \textbf{U5}) sentiments. Recently, reflexion provides a way to solve factual hallucination in LLMs. Therefore, a better-behaved approach should consider introducing the trusted learning to further alleviate the factual hallucination challenges of LLMs, thereby enhancing the credibility of LLMs in ASU.

To tackle the aforementioned challenges, we propose a \textbf{T}rusted \textbf{S}elf-reflexion \textbf{A}pproach (TSA) to our ChatASU task. Specifically, we firstly design a chat-style dialogue instruction to input into the LLMs, generating corresponding outputs. Then, we introduce an Aspect Chain Reasoning (ACR) task as an auxiliary task to boost the performance of the primary ASU task, which address the aspect coreference issue. Furthermore, we integrates trusted learning into reflexion mechanisms to alleviate the factual hallucination problem, thereby enhancing the ability of LLMs in understanding aspect sentiments within interactive scenarios. Finally, we employ a reinforcement learning strategy to optimize predictions. Detailed evaluations demonstrate the effectiveness of our proposed TSA. The main contributions of our work are summarized as follows:
\begin{itemize}
    \item We propose a new ChatASU task with a specially-designed ACR sub-task to address the coreference issue of aspects in dialogue ASU scenarios, which may open up a promising avenue for research in this direction.
    \item We incorporate both reflexion mechanisms and trusted learning for better understanding aspect chain and alleviating hallucinations problems, thereby enhancing the ability and credibility of LLMs in understanding aspect sentiments. 
    \item We meticulously annotate a high-quality Chinese dataset ChatASU to evaluate the aspect sentiments comprehension ability of LLMs within dialogue ASU scenarios. Our work marks the first of its kind, shedding light on coreference issue in dialogue ASU scenarios and contributing to the evaluation and enhancement of LLMs' performance.
\end{itemize}

\section{Related Work}
\begin{table*}[t]
\centering
\setlength{\abovecaptionskip}{0.5 ex}
\setlength{\belowcaptionskip}{-3 ex}
    \renewcommand{\arraystretch}{1.2}
	\addtolength{\tabcolsep}{5pt}
\setlength{\tabcolsep}{1.4mm}{
\begin{tabular}{ c c c   c c c c c c c}
\hline  
\multirow{2}{*}{{Split}}&\multirow{2}{*}{{\#Utterances(Dialogues)}}&\multirow{2}{*}{{\#Explicit}}&\multirow{2}{*}{{\#Implicit}}&\multicolumn{2}{c}{Aspect Chain}&\multicolumn{4}{c}{Quadruple} \\
\cmidrule(r){5-6}
\cmidrule(r){7-10}
~ &&&&{\#Max}&{\#Avg}& {\#Pos}&{\#Neu}&{\#Neg}& {\#Total} \\
\hline 
Train&21612(2400)&8959&6172&11&2.40&7234&472&1261&8967 \\
Valid&2727(300)&1161&770&8&2.45&894&57&180&1131 \\
Test&2723(300)&1146&733&9&2.46&987&71&144&1202 \\
\hline 
\end{tabular}
}
\caption{The statistics for our annotated ChatASU Dataset. \#Explicit denotes the number of explicit aspect entities. \#Implicit denotes the number of references towards explicit aspects (e.g., the reference ``\emph{this movie}'' for the explicit aspect ``\emph{WenChaorong}'' in Figure \ref{fig:introduction}). \#Max and \#Avg denote the max length and average length of the aspect chain.}
\label{tab:annocation} 
\end{table*}
\subsection{Aspect Sentiment Understanding}
Aspect Sentiment Understanding (ASU) is a fine-grained sentiment analysis task, which focuses on extracting sentiment information towards specific aspects within the text. Traditional ASU tasks focus on non-interactive scenarios, such as comment text~\cite{aste,asqp,chenxiao,emnlp}. In recent years, some studies observe the shortcomings posed by non-interactive scenarios and propose ASU tasks based on interactive scenarios, such as Question-Answering scenarios~\cite{QAWang}, dialogue scenarios~\cite{diaasq,casa}, while these studies focus on leveraging pre-trained language models (PLMs). Recently, the emergence of LLMs provides a new paradigm for NLP, which inspires us to explore the capabilities of LLMs in dialogue scenarios. Despite these studies exploring the ASU task, they ignore the issue of coreference, even though this issue is very ubiquitous in dialogue. 

Different from the above studies, we propose a new ChatASU task to evaluate the capability of LLM on coreference issue and construct a new dataset to address the coreference issue in dialogues. To our best knowledge, for the ASU task, we are the first study to address coreference issue in dialogue scenarios. In addition, we are devoted to exploring the ability of LLMs to understand dialogues.


\subsection{Reflexion Mechanism}
Recently, large language models (LLMs) have made a significant impact on various tasks. Although LLMs currently understand the language well, it currently suffers from the hallucination problem~\cite{huanjue1,huanjue2}. 
The majority of current studies use reflexion mechanisms to address hallucination problem, such as obtaining the inference path of the LLMs~\cite{cots}, performing actions through observed results~\cite{react}, searching for problems using a tree structure~\cite{tot}, model editing~\cite{edit}, and using heuristic rules to allow the model to reflect~\cite{reflexion}, etc., where \citet{reflexion} inspire our approach.

Different from the above studies, we propose a new Trusted Self-reflexion Approach (TSA) to ChatASU task, which is the first to integrate trusted learning into reflexion mechanisms to alleviate the LLMs-intrinsic factual hallucination problem.

\section{Dataset Construction for ChatASU}
\label{sec:dataset}
\begin{figure*}
\setlength{\abovecaptionskip}{0.5 ex}
\setlength{\belowcaptionskip}{-3 ex}
\begin{center}
    \subfloat{
 \includegraphics[scale=0.18]{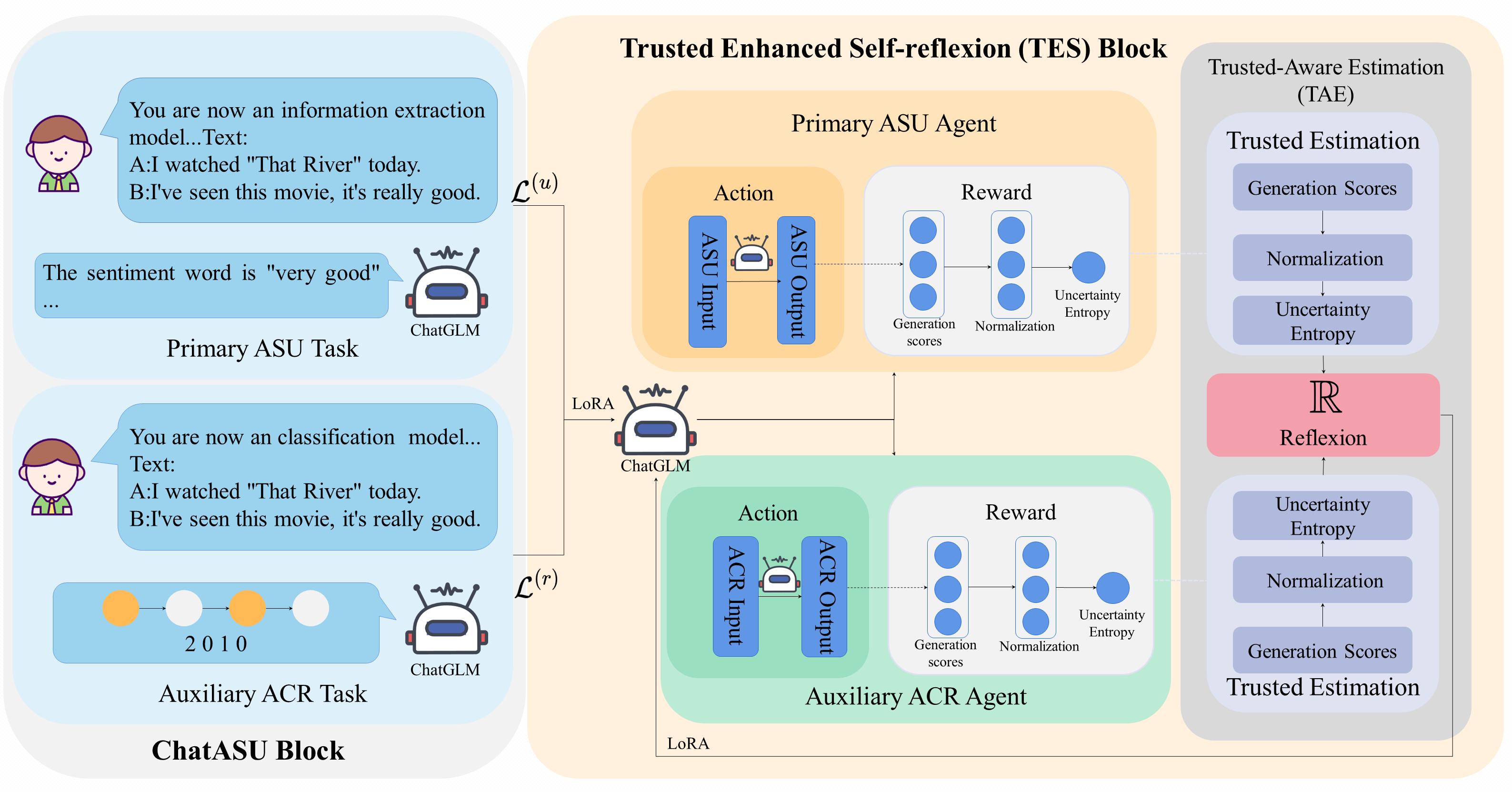}}
\caption{The overall framework of our Trusted Self-reflexion Approach (TSA), consisting of ChatASU Block and Trusted Enhanced Self-reflexion (TES) Block.}
\label{fig:model}
\end{center}
\end{figure*}

In order to evaluate the efficiency of the Trusted Self-reflexion Approach (TSA), we construct a new ChatASU dataset based on CASA~\cite{casa}, which consists of 3000 Chinese dialogues. Since previous studies ignore the coreference issue, this paper defines a new quadruple \textbf{``[Explicit Aspect, Implicit Aspect, Opinion, Polarity]''}, which uses explicit aspect and implicit aspect to consider coreference issue in dialogues. Different from existing annotation specifications~\cite{diaasq},  this paper does not annotate fine-grained aspect attributes or terms, which can significantly reduce the amount of annotation, and is easy to promote large-scale annotations and applications. It is worth noting that some studies have taken into account coreference between aspects~\cite{shengao1}, but these studies have not specifically targeted dialogue scenarios. In the following, we will introduce the annotation of explicit aspect, implicit aspect, opinion and polarity, respectively.


\textbf{Explicit Aspect} is used to integrate with \textbf{Implicit Aspect} to address the coreference issue. Specifically, we annotate the explicit aspect inside each dialogue based on the following two guidelines.

\textbf{(1)} To simulate a real dialogue environment, if an opinion appears in a sentence, we annotate the most specific aspect entity before current sentence as the explicit aspect. As shown in Figure \ref{fig:introduction} \textbf{U5}, the opinion phrase "bad reputation" appears and the opinion points to the aspect ``\emph{Wen Chaorong}'' in \textbf{U3}. ``\emph{Wen Chaorong}'' has a coreference expression ``\emph{a movie}'', but ``\emph{Wen Chaorong}'' is the most specific aspect, so we annotate ``\emph{Wen Chaorong}'' as an explicit aspect rather than ``\emph{a movie}''.

\textbf{(2)} If an aspect has no opinion expression, we do not annotate it. As shown in Figure \ref{fig:introduction} \textbf{U2}, for ``\emph{a relative}'' there is no expression of sentiment, we do not annotate this aspect.

\textbf{For Implicit Aspect}, we annotate the implicit aspect inside each dialogue based on the following two guidelines.

\textbf{(1)} If an aspect is pronoun of an explicit aspect, we annotate this aspect as an implicit aspect. As shown in Figure \ref{fig:introduction} \textbf{U5}, ``\emph{this movie}'' is a coreference of the explicit aspect ''Wen Chaorong'' but not the most specific aspect. Therefore, we annotate ``\emph{this movie}'' as an implicit aspect.

\textbf{(2)} If a more specific aspect appears after an explicit aspect, we do not modify  the previous aspect as an implicit aspect. As shown in \textbf{E1} and \textbf{E2}, ``\emph{Simon Pegg}'' is more specific compared to ``\emph{her}'', while ``\emph{her}'' comes before ``\emph{Simon Pegg}'', thus we don't modify ``\emph{her}'' to be an implicit aspect. \\
\textbf{E1} \textit{I like \underline{her} very much.}\\
\textbf{E2} \textit{Her name is \underline{Simon Pegg}.}

By combining the explicit and implicit aspects, we construct the aspect chain and introduce the ACR task. Specifically, an example of aspect chain  (``\emph{Wen Chaorong} $\rightarrow$ ... $\rightarrow$\emph{this movie}'') is shown in Figure \ref{fig:introduction}.

\textbf{Opinion and Polarity} is used as the sentiment annotation for the aspect. Specifically, we annotate the opinion and polarity inside each dialogue based on the following three guidelines.

\textbf{(1)} We annotate words or phrases that express explicit sentiment. As shown in Figure \ref{fig:introduction} \textbf{U5}, ``\emph{bad reputation}'' and ``\emph{pretty good}'' have explicit sentimental  expressions. Therefore, we annotate ``\emph{bad reputation}'' and ``\emph{pretty good}'' as opinions and annotate their sentiment polarities as ``\emph{negative}'' and ``\emph{positive}'', respectively.

\textbf{(2)} We do not annotate words or phrases with weak sentiment expressions. As shown in Figure \ref{fig:introduction} \textbf{U4}, the phrase ``\emph{you recommended it should be good to see}'' implies a positive sentiment, while the sentiment expression is not strong enough, thus we do not annotate it.

\textbf{(3)} We categorize opinion as ``\emph{positive}'', ``\emph{negative}'', and ``\emph{neutral}''  based on their sentiment orientation.


During the annotation process, we employ 10 professional annotators. Each dialogue is annotated by two annotators, if they are in disagreement on the annotation result of a dialogue, we employ an extra domain expert to make the final decision. Finally, we randomly split the dataset into training, validation, and test sets in a ratio of 8:1:1, and the statistics of  the dataset are shown in Table \ref{tab:annocation}. Besides, following \citet{biaozhu}, we use the matching F1 score and accuracy between two annotators as measures of annotation consistency for the ChatASU dataset. The final F1 score and accuracy are 86.76 and 83.16, respectively.

\section{Trusted Self-reflexion Approach}
In this section, we formulate the Chat-based Aspect Sentiment Understanding (ChatASU) task, which is consist of  two sub-tasks: the primary Aspect Sentiment Understandin (ASU) task and the auxiliary  Aspect Chain Reasoning (ACR) task. The specific formulation of the two sub-tasks is described in section 4.1. 
In this paper, we propose a Trusted Self-reflexion Approach (TSA) for the ChatASU task. The TSA utilizes the ACR task as an auxiliary task to enhance the performance of the ASU task. Additionally, TSA integrates trusted learning into reflexion mechanisms to alleviate the LLMs-intrinsic factual hallucination problem in the ChatASU task. Figure \ref{fig:model} shows the overall architecture of our approach, which is consist of two major parts: \textbf{1)} ChatASU Block. \textbf{2)} Trusted Enhanced Self-reflexion (TES) Block.

\subsection{ChatASU Block}
\textbf{Backbone LLM.} A lot of LLMs are open-sourced currently, we consider using ChatGLM-6B~\cite{chatglm} as the backbone. ChatGLM is optimized for Chinese Q\&A and dialogues, endowing it with strong Chinese language comprehension abilities. Therefore, we utilize ChatGLM as the backbone in the ChatASU task to explore the capabilities of LLMs in addressing coreference issue.

The ChatASU task is divided into two sub-tasks, the Primary ASU Task and the Auxiliary ACR Task. Their specific formulations are as follows.

\textbf{Primary ASU Task.} Given a dialogue $ \bm{{\rm C}} = \{\bm{{ c}}_1,\bm{{c}}_2,...,\bm{{c}}_n\}$, where $\bm{{c}}_i$ represents the $\bm{{i}}$-th utterance and $n$ represents the total number of utterances. The ASU task aims to identify all the quadruples $ (\bm{{e}}_i, \bm{{r}}_i, \bm{{o}}_i, \bm{{p}}_i)$, where $\bm{{o}}_i$ represents opinion in $\bm{{c}}_i$ and $\bm{{e}}_i$ is the most specific explicit aspect of $\bm{{o}}_i$ in $\bm{{c}}_j$, $j \in \{1,2,...,i\}$. $\bm{{p}}_i$ represents the sentiment polarity of $\bm{{o}}_i$. Specifically, we need to identify the implicit aspect $\bm{{r}}_i$ based on $\bm{{e}}_i$ in $\bm{{c}}_i$. When no implicit aspect is present, we label it as $\bm{{null}}$.

In this paper, we employ the chat-style to fine-tune ChatGLM, enabling ChatGLM to extract quadruples effectively. Inspired by instruct learning~\cite{instructlearning}, we add an ``instruct'' statement at the beginning of the input sequence and utilize a question-and-answer format to obtain outputs, from which we extract the quadruple. The specific process is described as follows.

$\bullet$ \textbf{$\bm{{\rm Input_{ASU}}}$.} We first present instructions to TSA, which are used to give a clear definition of the task. The \textit{Instruction} is formulated as follows: 
\emph{``You are now an information extraction model. Please help me to extract opinions from the input and tell me the sentiment polarity of the opinions, what the explicit aspect referred to by the opinion is, and what pronoun is used for the explicit aspect in the utterance where the opinion occurs.''}

The input for ChatGLM in the ASU task is obtained as follows.
\textbf{$\bm{{\rm Input_{ASU}}}$} = \textit{Instruction $[ \cdot ]$ text }, where $[\cdot]$ represents string splicing operation, and \textit{text} is the dialogue text. 

$\bullet$ \textbf{$\bm{{\rm Output_{ASU}}}$.} For the output of our approach, we use the way of chat-style like ChatGPT to obtain it. Taking the text ``\emph{I watched That River today, the movie is very good to recommend you to see}'' as an example, the output format is as follows. \textbf{$\bm{{\rm Output_{ASU}}}$} = \textit{The opinion is ``very good''. The sentiment tendency is ``POS''. The opinion refers to the explicit aspect ``That River''. The pronoun of ``That River'' is ``the movie''.} After getting the output, we filter it to get the target quadruple (\textit{That River},\textit{the movie},\textit{very good},\textit{POS}).

\textbf{Auxiliary ACR Task.} Given an explicit aspect set $\bm{{\rm E}}=\{\bm{{e}}_1,\bm{{e}}_2,...,\bm{{e}}_m\}$, where $\bm{{e}}_i$ represents the $\bm{i}$-th explicit aspect, and $m$ represents the number of explicit aspect in the dialogue. If an utterance contains an explicit aspect or implicit aspect, we label it as 2 or 1, respectively. Otherwise the label is 0. Particularly, if both explicit aspect as well as implicit aspect occur in an utterance, we label it as 2. As shown in Figure \ref{fig:model}, the output of ACR task is [2, 0, 1, 0] for a given aspect, which represents that the explicit aspect exists in the first utterance, the implicit aspect exists in the third utterance and no coreference in the second and fourth utterances. 


In this paper, we use the chat-style to handle the ACR task and obtain the aspect chain. The input and output formats for the ACR task are as follows.

$\bullet$ \textbf{$\bm{{\rm Input_{ACR}}}$.} The input of the ACR task also concatenates instruction and text. The format of \textit{Instruction} is as follows:
\textit{``You are now an classification model to judge which utterance in this dialogue appears to be the coreference of $\bm{{e}}_i$, outputs 2 if it is an explicit aspect, 1 if it is an implicit aspect, and otherwise 0. Output a sequence of 0, 1, and 2, the length of which is the number of dialogues.''} 

The input for ChatGLM in the ACR task is obtained as follows.
\textbf{$\bm{{\rm Input_{ACR}}}$} = \textit{Instruction $[ \cdot ]$ text }, where $[\cdot]$ represents string splicing operation, and \textit{text} is the dialogue text.

$\bullet$ \textbf{$\bm{{\rm Output_{ACR}}}$.} The output of the ACR task is a sequence consisting of 0, 1 and 2, with the length of the sequence equal to the number of utterances in the dialogue.

\subsection{Trusted Enhanced Self-reflexion}
Trusted Enhanced Self-reflexion (TES) block integrates trusted learning into reflexion mechanisms by reinforcement learning to alleviate the factual hallucination problem. The TES block comprises two agents: the Primary ASU Agent and the Auxiliary ACR Agent.  The specific descriptions of these two agents are as follows.

\textbf{Primary ASU Agent} integrates trusted learning into ASU task to alleviate the factual hallucination problem. The process of ASU Agent in reinforcement learning is as follows. In state $s_t$, where $t$ represents the $t$-th time step, we execute action $a_t$ according to policy $\pi(a_t | s_t)$. The specific action and reward of ASU agent are as follows.

$\bullet$ \textbf{Action.} We get the ASU task output of ChatGLM. The action is formulated as follows.
\begin{equation}
\textbf{$\bm{{\rm Output_{ASU}}}$} = {\rm ChatGLM}(\textbf{$\bm{{\rm Input_{ASU}}}$})
\end{equation}

$\bullet$ \textbf{Reward.} We get the generation scores $\bm{{\rm Score_{ASU}}}$ from the \textbf{$\bm{{\rm Output_{ASU}}}$}, where the generation scores $\bm{{\rm Score_{ASU}}}$ represent the path scores of ChatGLM's beam search in the ASU task. The reward formula for the ASU agent is as follows.
\begin{equation}
\mathbb{R}_{ASU} = \mathbb{R}_{TE}(\bm{{\rm Score_{ASU}}})
\end{equation} 
where $\mathbb{R}_{TE}(\cdot)$ is described in Eq.(6).

\textbf{Auxiliary ACR Agent} integrates trusted learning into ACR task to alleviate the factual hallucination problem. The process of ACR agent in reinforcement learning is same as ASU Agent. The specific action and reward of ACR agent are as follows.

$\bullet$ \textbf{Action.} We get the ACR task output of ChatGLM. The action is formulated as follows.
\begin{equation}
\textbf{$\bm{{\rm Output_{ACR}}}$} = {\rm ChatGLM}(\textbf{$\bm{{\rm Input_{ACR}}}$})
\end{equation} 

$\bullet$ \textbf{Reward.} We get the generation scores  $\bm{{\rm Score_{ACR}}}$ from the \textbf{$\bm{{\rm Output_{ACR}}}$}, where the generation scores $\bm{{\rm Score_{ACR}}}$ represent the path scores of ChatGLM's beam search in the ACR task. The reward formula for the ACR agent is as follows.
\begin{equation}
\mathbb{R}_{ACR} = \mathbb{R}_{TE}(\bm{{\rm Score_{ACR}}})
\end{equation}


\textbf{Trusted-Aware Estimation (TAE)} generates rewards through trusted learning and reflexion mechanisms, thereby motivating the ChatGLM to produce credible results. The TAE Block is consist of two parts, Trusted Estimation and Trusted Reflexion.

$\bullet$ \textbf{Trusted Estimation (TE)} utilizes the difference in generation scores obtained from beam search as a measure of confidence, encouraging ChatGLM to produce trustworthy results. Specifically, we obtain the output of ChatGLM through beam search and obtain generation scores, denoted as $\bm{{\rm G}} = \{\bm{g}_1, \bm{g}_2, ..., \bm{g}_n\}$, where $n$ represents the number of generation scores. Next, we enhance the data by normalizing these generation scores.
\begin{equation}
\bm{\hat{g}}_i \! \! = \! \!  \bm{{\rm Normalization(G)}} \! \! = \! \!  \frac{\bm{g}_i \! \! - \! \! min(\bm{{\rm G}})}{max(\bm{{\rm G}}) \! \! - \! \! min(\bm{{\rm G}})}
\end{equation}
where $\bm{g}_i$ represents the $\bm{i}$-th generated score, $i = 1, 2, ..., n$. $max(\cdot)$ and $min(\cdot)$ represent  the operation of taking the maximum value and the minimum value, respectively.

After obtaining the enhanced results, we use a reward function to calculate the reward. The formal formula for the reward function is as follows.
\begin{equation}
\mathbb{R}_{TE}(\hat{G}) = - \sum_{j=1}^{m} \frac{1}{ \sum_{i=1}^{n} \bm{m \hat{g}}_i \log{\bm{\hat{g}}_i}}
\end{equation}
where $\hat{G} = \{\bm{\hat{g}}_1, \bm{\hat{g}}_2, ..., \bm{\hat{g}}_n \}$. $n$ represents the number of generated scores in an output, and $m$ represents the number of outputs. 

As shown in Eq.(6), the reward function is based on the entropy function, but it differs from traditional entropy. In traditional entropy, higher entropy indicates greater disorder, i.e., higher uncertainty and less reliability in the results. However, our reward function operates in the opposite way. When the reward is larger, it means that there is a greater difference in generation scores. We consider that ChatGLM's generation results have smaller uncertainty. In this context, ChatGLM is more confident in the generated results, making the generated outcomes more reliable.

$\bullet$ \textbf{Trusted Reflexion} is the final reward function. Inspired by \cite{reflexion}, we penalize the ChatGLM when it performs repetition to generate the same result. The final reward integrates trusted learning into reflexion mechanisms to alleviate the LLMs-intrinsic factual hallucination problem. The final reward is formulated as follows.
\begin{equation}
\mathbb{R} \! \! = \! \! \begin{cases} 
\alpha \mathbb{R}_{ACR} \! \! + \! \! \beta \mathbb{R}_{ASU} \! \! + \! \! \gamma (\mathbb{R}_{Rp} \! \! + \! \! \mathbb{R}_{Ra}),\! \! & \! \! \mbox{if }\bm{{p}} \! \! = \! \! 0 \\
\alpha \mathbb{R}_{ACR} \! \! - \! \! \beta \bm{{p}} \! \! + \! \! \gamma (\mathbb{R}_{Rp} \! \! + \! \! \mathbb{R}_{Ra}),\! \! & \! \! \mbox{else }
\end{cases}
\end{equation}
where $\bm{{p}}$ is the number of repeat generation, and $\alpha$, $\beta$, $\gamma$ are hyper-parameters. In reinforcement learning, models often suffer from catastrophic forgetting during training~\cite{yiwang}. Following \citet{wangg}, we introduce the F1 score of ASU task and ACR task (i.e., $\mathbb{R}_{Rp}$ and $\mathbb{R}_{Ra}$) as rewards. The formulas for these two rewards are as follows.
\begin{equation}
\mathbb{R}_{Rp}   =  F1 = \rm \frac{2 \cdot Pr_{asu}  Re_{asu}}{Pr_{asu} + Re_{asu}}
\end{equation}
\begin{equation}
\mathbb{R}_{Ra}   =  F1 = \rm \frac{2 \cdot Pr_{acr} Re_{acr}}{Pr_{acr} + Re_{acr}}
\end{equation}
where $\rm {Pr} = \frac{N_{cp}}{N_{t}}$ and $\rm Re = \frac{N_{cp}}{N_{p}}$. $N_{cp}$ represents the number of correct predictions. $N_{t}$ and $N_{p}$ represent the number of quadruple in label and the number of quadruple in predict, respectively.

\begin{table*}[t]
\setlength{\abovecaptionskip}{0.5 ex}
\setlength{\belowcaptionskip}{-3 ex}
    \renewcommand{\arraystretch}{1.2}
	\addtolength{\tabcolsep}{5pt}
\begin{center}
\setlength{\tabcolsep}{0.85mm}{
\begin{tabular}{c|l|c c c c c c c c }
\toprule[1.2pt] 
&\multirow{2}{*}{Approach}&\multicolumn{4}{c}{Single}&\multicolumn{3}{c}{Pair}& Quadruple \\
 \cmidrule(r){3-6}  \cmidrule(r){7-9}  
~ & & Explicit & Implicit & Opinion & Polarity & E-O & E-I & I-O & Extraction \\
\hline
\multirow{2}{*}{\textbf{PLM}}&ASQP\cite{asqp}&73.08&60.36&61.35&81.52&49.88&50.45&44.59&36.66 \\
~&DiaASQ\cite{diaasq}&49.42&41.7&43.24&56.11&38.61&36.78&34.23&29.09 \\
\hline
\multirow{7}{*}{\textbf{LLM}}&ChatGPT(zero-shot)&47.65&55.6&41.88&64.99&27.43&31.05&26.71&22.38 \\
&ChatGPT(In-context learning)&47.82&56.52&43.47&69.57&37.13&34.78&43.48&30.43 \\

&ChatGLM\cite{chatglm}&67.49&73.70&61.70&82.21&51.32&57.44&52.00&43.49 \\
~& 
Reflexion\cite{reflexion}&68.84&74.11&65.37&86.59&53.32&57.61&53.98&44.62 \\
~&\textbf{TSA}&\textbf{70.58}&74.45&\textbf{65.98}&\textbf{86.85}&\textbf{55.05}&\textbf{58.92}&\textbf{54.81}&\textbf{46.34} \\
~& - w/o Trusted Learning &69.84&74.22&64.88&86.59&53.66&57.8&53.50&44.88  \\
~& - w/o ACR Task & 68.98&\textbf{75.59}&64.90&86.45&53.14&58.45&54.53&44.98 \\

\bottomrule[1.2pt]
\end{tabular}
}
\caption{Comparison of several state-of-the-art approaches on ASU task, where ``Single'' denotes the F1 score extracted separately for each element inside quadruple and ``Pair'' denotes the F1 score for a pair of two elements inside quadruple. E, I, O denote explicit aspect, implicit aspect, opinion, respectively.}
\label{tab:results} 
\end{center}
\end{table*}

\subsection{Optimization for TSA} 
We fine-tune ChatGLM on the ASU task and ACR task using cross-entropy loss. The loss functions for these two tasks are as follows.
\begin{equation}
\setlength\abovedisplayskip{3pt}
\setlength\belowdisplayskip{3pt}
     \mathcal{L}^{(u)} = -\sum_{i=1}^N \sum_{j=1}^K y_{ij}\mathrm{log}(\hat{y}_{ij})
     \label{eq:cross-entorpy}
\end{equation}
\begin{equation}
\setlength\abovedisplayskip{3pt}
\setlength\belowdisplayskip{3pt}
\begin{split}
  \mathcal{L}^{(r)} = -\sum_{i=1}^N \sum_{j=1}^K w_{ij}\mathrm{log}(\hat{w}_{ij})
\end{split}
\end{equation}
where $y$ and $\hat{y}$ represent the labels and the prediction in the ASU task, respectively.  $w$ and $\hat{w}$ represent the labels and the prediction in the ACR task, respectively. $N$ and $K$ represent the length of the label and the length of the vocabulary, respectively. $\mathcal{L}^{(u)}$ and $\mathcal{L}^{(r)}$ represent the loss function of the ASU task and ACR task, respectively. The total loss function is $\mathcal{L} \!=\! \mathcal{L}^{(u)} + \mathcal{L}^{(r)}$.

In the reinforcement learning part, we use the Proximal Policy Optimization (PPO)~\cite{ppo} algorithm to maximize our reward. The objective function for the reinforcement learning algorithm is as follows.
\begin{equation}
\setlength\abovedisplayskip{3pt}
\setlength\belowdisplayskip{3pt}
\max \limits_{\theta} J(\theta)  \! \!=\! \! \max \limits_{\theta} \sum_{\tau} P(\tau ; \theta) R(\tau)
\end{equation}
where $\tau = (s_1,a_1,...,s_T,a_T)$ is the state-action trajectory, $R(\cdot)$ is the reward function and $\theta$ is the parameter of the network.

\section{Experiments}
\subsection{Experimental Settings and Metrics}
We empirically evaluate the performance of the TSA on the ASU task using the ChatASU dataset, which is described in Section \ref{sec:dataset}.

We fine-tune ChatGLM-6B\footnote{https://cloud.tsinghua.edu.cn/d/fb9f16d6dc8f482596c2/} using LoRA~\cite{lora}\footnote{https://github.com/liucongg/ChatGLM-Finetuning}. We set the dimension, scaling factor, dropout rate of the LoRA matrix to be 8, 32, 0.1, respectively, while keeping other parameters at their default values.
During fine-tuning, we utilize the Adam optimizer with a learning rate of 2e-4 and weight decay of 5e-4. ChatGLM is trained for 5 epochs on a single A100 40G GPU with a batch size of 2.  Training is conducted using Deepspeed\footnote{https://github.com/microsoft/DeepSpeed}. In the reinforcement learning part, the learning rate of ChatGLM is 1e-5 and the batch size is 1. The hyper-parameters $\alpha$, $\beta$ and $\gamma$ are 15, 5 and 3, respectively. The model requires approximately four hours for one training session.

We measure our approach through three perspectives. 1) Single: individual extract of each element. 2) Pair: extract the element pair, i.e., explicit aspect-implicit aspect pair, explicit aspect-opinion pair, implicit aspect-opinion pair. 3) Quadruple Extraction: extract the complete quadruple. Following the prior works~\cite{diaasq}, the performance is evaluated with Macro-F1. Moreover, t-test is used to evaluate the significance of the performance difference~\cite{ttest}.
\subsection{Baselines}
Models like ChatGLM have a much larger number of parameters compared to models like T5. Comparing ChatGLM with models like T5 is unfair. Therefore, we categorize baselines into methods based on Pre-trained Language Models (PLMs) and methods based on Large Language Models (LLMs) according to the number of parameters. \textbf{For PLMs}, ASQP~\cite{asqp} utilizes the T5~\cite{t5} to get the output via paraphrasing. DiaASQ~\cite{diaasq} introduces attention mask matrices and combines RoBERTa to model dialogue-specific features via RoBERTa~\cite{Droberta}. Following \citealt{toward}, we use the traditional approach to extract quadruples for these PLMs baselines. \textbf{For LLMs}, in ChatGPT, we randomly choose 30 samples and use two ways (zero-shot and in-context learning) to evaluate the capabilities of the ASU task. In zero-shot, we follow the method in \citet{chatie} to obtain the results. In in-context learning, we follow the method in \citet{icl} and provide a sample for each input to obtain the results. ChatGLM~\cite{chatglm} directly extracts quadruples through fine-tuning. Reflexion~\cite{reflexion} allows the model to self-reflect when its generated results exceed the threshold by setting a threshold value, and this approach is the state-of-the-art approach in the ASU task.

The hyper-parameters of these baselines reported by their public papers are still adopting the same setting, and the others are tuned according to the validation set. We reproduce the approach proposed in Reflexion~\cite{reflexion} with ChatGLM as the backbone. To facilitate the corresponding research in this direction, all codes together with datasets will be released via Github\footnote{https://github.com/Atend9/ChatASU}.
\begin{figure}
    \centering
    \begin{subfigure}[b]{0.2\textwidth}
    \includegraphics[width=\textwidth]{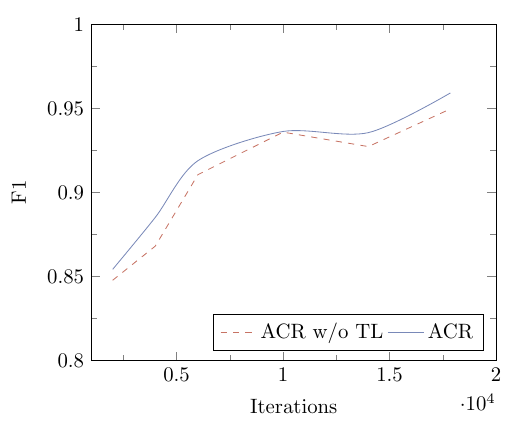}
    \caption{ACR}
    \label{fig:uncertainty}
    \end{subfigure}
    \hspace{0.03\textwidth}
    \begin{subfigure}[b]{0.2\textwidth}
    \includegraphics[width=\textwidth]{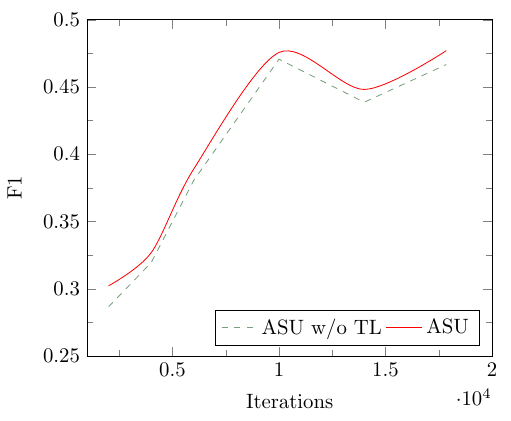}
    \caption{ASU}
    \label{fig:task}
    \end{subfigure}
    \setlength{\abovecaptionskip}{0.5 ex}
  \setlength{\belowcaptionskip}{-4 ex}
    \caption{ (a) The performance of our TSA to ACR task with or without TL during different training steps. (b) The performance of our TSA to ASU task with or without TL during different training steps.}
    \label{fig:task}
\end{figure}
\begin{figure*}
\begin{center}
\setlength{\abovecaptionskip}{0.5 ex}
\setlength{\belowcaptionskip}{-4 ex}
    \subfloat{
 \includegraphics[scale=0.205]{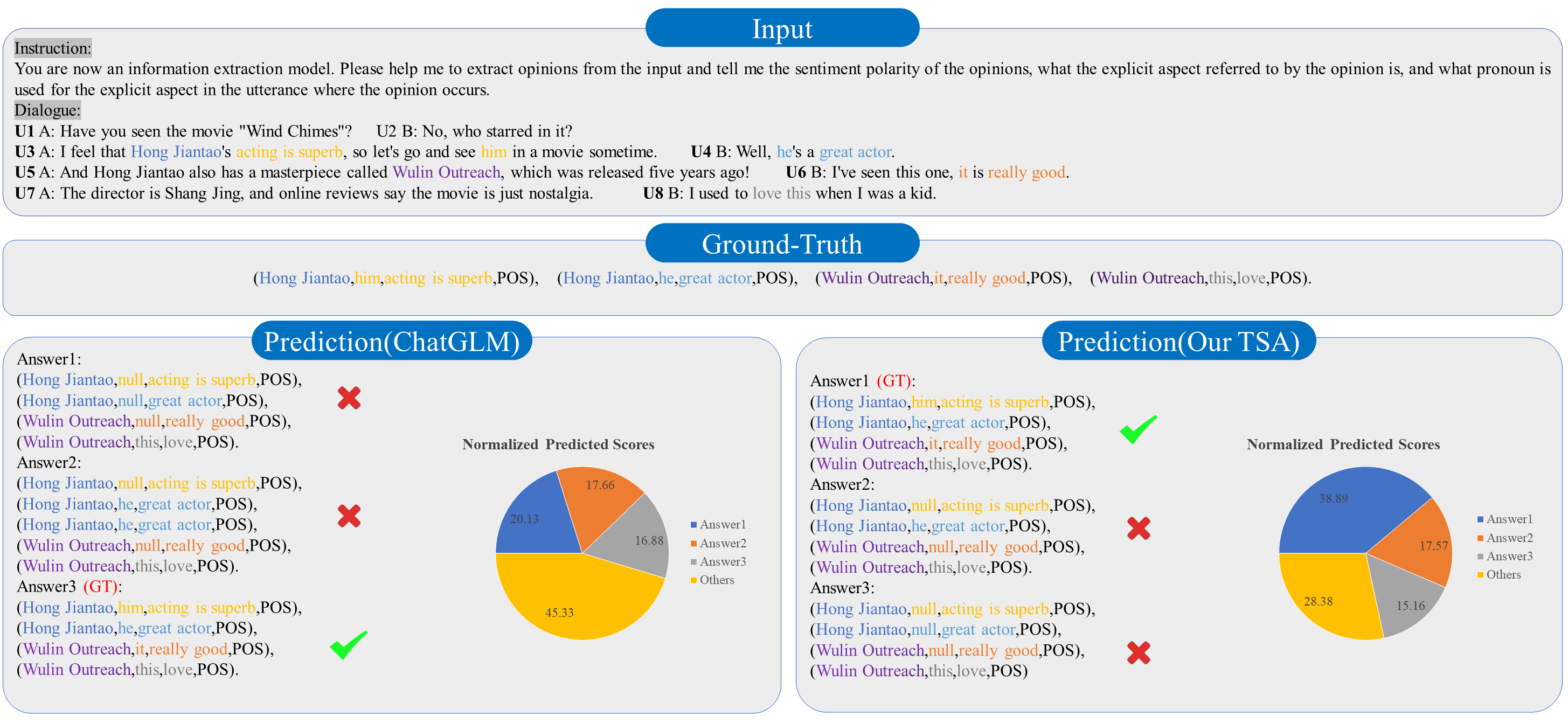}}
\caption{A dialogue example (eight utterances) with their four ground-truth quadruple from the test data of
ChatASU dataset. Normalized Predicted Scores denote the normalized predicted percentage for top-3 and ``other'' generated answers. GT denotes the ground-truth. }  
\label{fig:analysis}
\end{center}
\end{figure*}
\subsection{Experimental Results}
Table \ref{tab:results} shows the performance of different approaches to ASU task. From this table, we can see that \textbf{1)} The approaches based on LLMs (i.e., Reflexion, TSA) outperform the approaches based on PLMs (i.e., ASQP with T5, DiaASQ with RoBERTa). This justifies the powerful comprehension of language by the LLMs. In ChatGPT, due to the model is not open-sourced and only available for interactive use, it results in unsatisfactory performance on ASU tasks.
\textbf{2)} The models designed for other ASU-related tasks do not perform well in the ASU task. Both ASQP and DiaASQ, based on PLMs and not taking into account the coreference issue, exhibit lower performance than TSA on the ASU task.
\textbf{3)} The TSA outperforms other LLM-based approaches on the ASU task. Specifically, the TSA outperforms Reflexion in Single, Pair, and Quadruple Extraction by 0.73\%, 1.29\% and 1.72\% on ASU task, respectively. Significance test shows that these improvements are all significant (p-value < 0.05). Particularly, the TSA outperforms GhatGLM in Single, Pair, and Quadruple Extraction by 3.19\%, 2.67\%, and 2.85\% on ASU task, respectively. This justifies the effectiveness of TSA.

To further illustrate the effectiveness of the TSA, we analyze the impact of each part in TSA. Table \ref{tab:results} shows that there is a decrease in our approach performance without the trusted learning and the ACR task. \textbf{1)} w/o Trusted Learning (i.e., removing Eq.(5) and Eq.(6)). As shown in Table \ref{tab:results}, the performance of TSA in Quadruple Extraction decreases by 1.46\%. This indicates that trusted learning is effective in alleviating the problem of factual hallucination.
\textbf{2)} w/o ACR Task. As shown in Table \ref{tab:results}, TSA experiences a decrease in extraction performance by 1.14\% in implicit aspect, but it shows an improvement of 1.36\% in pair extraction.  This indicates that TSA has better relation extraction capabilities in ChatASU and also illustrates the effectiveness of ACR tasks in addressing coreference issues. These further justify the effectiveness of TSA, which again encourages us to consider coreference issue and factual hallucination problem in the ASU task.

\section{Analysis and Discussions}
\subsection{Importance and Robustness Study for ACR Task and Trusted Learning}
To verify the importance and robustness of ACR task and trusted learning, we visualize the F1 scores on the ASU task during the training stage of ChatGLM. As shown in Figure \ref{fig:task}, we can see that 1) The F1 score of the ACR task grows with the training process, indicating that the ACR task can be efficiently learned by the ChatGLM to address the coreference issue. 2) During the training process, the ACR task and ASU task exhibit similar growth trends. Particularly, between step 15000 and 18000, there is a noticeable improvement in performance on both ACR and ASU tasks. This further validates that the ACR task effectively assists ChatGLM in enhancing its performance on the ASU task. 3) When trusted learning is incorporated into the training process of ChatGLM, there is a significant improvement in performance on both the ACR and ASU tasks. This demonstrates the effectiveness and robustness of trusted learning.


 \subsection{Qualitative Study}
We provide a qualitative analysis of TSA in ASU task on the ChatASU dataset. Figure \ref{fig:analysis} illustrates samples with coreference issue and factual hallucination problem, showcasing their prediction and normalized predicted scores. We select a dialogue example  from ChatASU dataset to analyze the coreference issue and factual hallucination problem. From Figure \ref{fig:analysis}, we can see that 1) ChatGLM fails to recognize the pronoun of ``\emph{Hong Jiantao}'' in the text, while TSA accurately identifies the pronoun ``\emph{him}'' referring to ``\emph{Hong Jiantao}'' in \textbf{U4} and ``\emph{he}'' referring to ``\emph{Hong Jiantao}'' in \textbf{U3}. This demonstrates that TSA can effectively address coreference issue, while ChatGLM exhibits coreference issue.
2) The normalized predicted scores of TSA exhibit significant differences, indicating higher confidence in the generated results. This suggests that the outputs generated by TSA are more reliable. However, in ChatGLM, the differences between the highest prediction scores for Answer1, Answer2, and Answer3 are very small. This implies that ChatGLM lacks confidence in its generated results, making them less reliable. However, TSA enhances the reliability of its generated results through trusted learning.

\section{Conclusion}
In this paper, in order to address the coreference issue and explore the LLMs’ ability in understanding aspect sentiments in dialogues, we introduce a ChatASU task with a specially-designed ACR sub-task and construct a high-quality human annotated dataset for ChatASU. Besides, LLMs currently suffer from the hallucination problem. With these in mind, we propose a Trusted Self-reflexion Approach (TSA), which integrates trusted learning into reflexion mechanisms to address the coreference issue and alleviate the hallucination problem. Detailed experiments demonstrate the effectiveness of TSA. In our future work, we would like to transfer our approach to mutimodal scenarios (e.g., multimodal aspect-based sentiment analysis) and introduce more information (e.g., eye-contact information) to address the coreference issue in mutimodal dialouges. Moreover, we would like to introduce more reflexion-based approaches (e.g., model editing) to further alleviate the hallucination problem.

\section*{Acknowledgements}
We thank our anonymous reviewers for their helpful comments. This work was supported by three NSFC grants, i.e., No.62006166, No.62376178 and No.62076175. This work was also supported by a Project Funded by the Priority Academic Program Development of Jiangsu Higher Education Institutions (PAPD).

\section{Bibliographical References}
\bibliographystyle{lrec-coling2024-natbib}
\bibliography{lrec-coling2024-example}

\bibliographystylelanguageresource{lrec-coling2024-natbib}
\bibliographylanguageresource{languageresource}

\end{document}